\RequirePackage{amsmath}         

\documentclass[runningheads,a4paper]{llncs}

\usepackage{graphicx}            
\usepackage{amssymb}             
\usepackage{amsfonts}            
\usepackage{enumitem}            
\usepackage{multirow}            
\usepackage{siunitx}             
\usepackage{url}                 
\usepackage{xspace}              
\usepackage[T1]{fontenc}         
\usepackage[hidelinks]{hyperref} 
\usepackage{wrapfig}
\usepackage{todonotes}
\usepackage{units}

\graphicspath{{images/}}
\DeclareGraphicsExtensions{.pdf,.png,.jpg,.jpeg}

\newcommand{\degreem}{^{\circ}} 

\newcommand{\seclabel}[1]{\label{sec:#1}}

\newcommand{\figlabel}[1]{\label{fig:#1}}

\newcommand{\secref}[1]{Section~\ref{sec:#1}\xspace}

\newcommand{\figref}[1]{Fig.~\ref{fig:#1}\xspace}

\newcommand{\nop}{NimbRo\protect\nobreakdash-OP\xspace}
\newcommand{\noptwo}{NimbRo\protect\nobreakdash-OP2\xspace}

\newcommand{\iguhop}{igus\textsuperscript{\tiny\circledR}$\!$ Humanoid Open Platform\xspace}

\newcommand{\degree}{$\degreem$\xspace}

\begin{document}

\mainmatter

\title{Grown-up NimbRo Robots Winning RoboCup 2017 Humanoid AdultSize Soccer Competitions}
\titlerunning{Grown-up NimbRo Robots Winning RoboCup 2017 Humanoid AdultSize}

\author{Grzegorz Ficht, Dmytro Pavlichenko, Philipp Allgeuer, Hafez Farazi, Diego Rodriguez, Andr\'{e} Brandenburger, Johannes K\"{u}rsch, Michael Schreiber and Sven Behnke}
\authorrunning{Ficht, Pavlichenko, Allgeuer, Farazi, et al.}

\institute{Autonomous Intelligent Systems, Computer Science, Univ.\ of Bonn, Germany\\
\url{{ficht, pavlichenko, pallgeuer, rodriguez, behnke}@ais.uni-bonn.de},
\url{http://ais.uni-bonn.de}}

\maketitle

\begin{abstract}
The ongoing evolution of the RoboCup Humanoid League led in 2017 to the introduction of one vs. one soccer games for the AdultSize robots, which motived our team NimbRo to enter this category. 
In this paper, we present the mechatronic design of our upgraded robot Copedo and the newly developed NimbRo-OP2, which received the RoboCup Design Award.
We also describe improved approaches to visual perception of the game situation, including compassless localization on a soccer field with symmetric appearance, and the generation of soccer behaviors.
At RoboCup 2017 in Nagoya, our robots played very well, winning the AdultSize soccer tournament with high scores. 
Our robots also won the technical challenges and we present the developed solutions.
\end{abstract}

\section{Introduction}

\begin{figure}[!b]
\parbox{\linewidth}{\centering
\includegraphics[height=41.5mm]{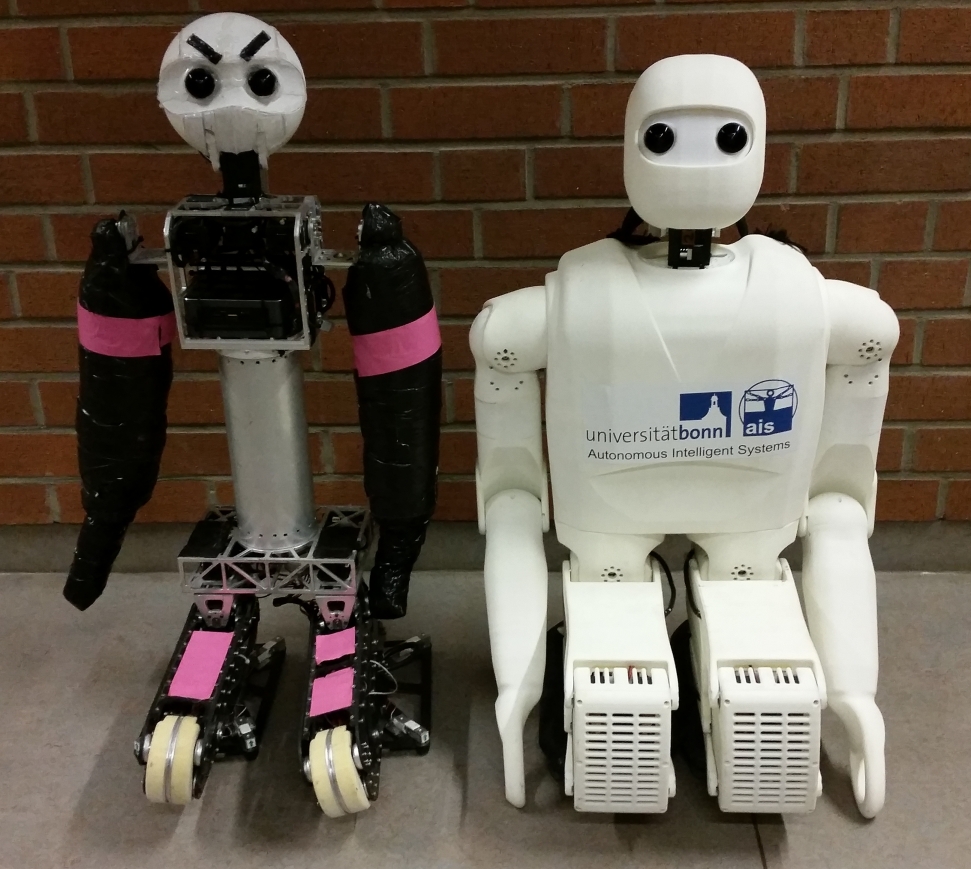}\hspace{0.005\linewidth}
\includegraphics[height=41.5mm]{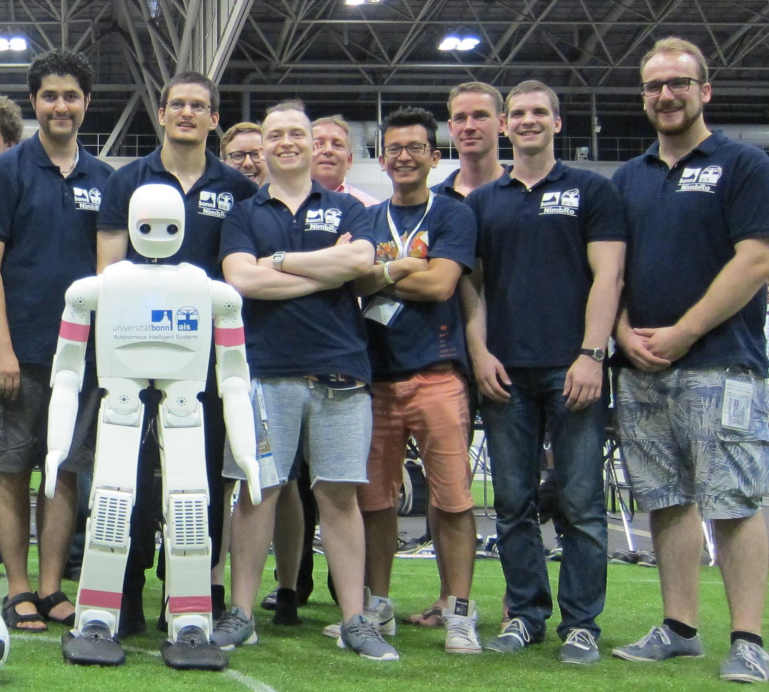}}
\caption{Left: Our AdultSize Robots Copedo and the NimbRo-OP2. Right: The NimbRo team at RoboCup 2017 in Nagoya, Japan.}
\figlabel{nimbro_team}
\end{figure}

With the ongoing evolution of its rules, RoboCup Humanoid League, competition each year is becoming more 
realistic and closer to playing games against human soccer players. In 2017, for the first time, AdultSize class robots were playing 
one vs. one matches, which meant bringing the soccer game to a larger scale, along with all of its inherent difficulties.
In previous years, our team NimbRo has competed in the Humanoid League within the KidSize, and more recently, the TeenSize classes~\cite{Lee:RoboCup2011,missura2013robocup,Missura:RoboCup2013,Farazi2017}.

In order to compete in the AdultSize category, it was necessary to procure hardware that is not only in compliance with the new set of rules,
but also performs well under the conditions that a robot may face during the competition. For that purpose, we have prepared 
two robots capable of play: Copedo and NimbRo-OP2. Our robust hardware played a major role in our success, where over the span of five 
games in the soccer tournament and a set of technical challenges, our robots secured victory in both, without the need for repairs or maintenance.
We describe the mechanic design of our robots in \secref{hardware_design}, and our approach for the technical challenges in \secref{technical_challenges}.

Our open-source software is advanced with each RoboCup, with many modules being added and existing ones being improved. With the 
introduction of identically looking field halves in 2015 and the ban on magnetometers in 2017, localisation has become much more difficult. The approach we 
used for breaking the symmetry and localising in these conditions, as well as how we combine this information with our ball position estimate to 
plan further actions are described in \secref{software_design}.
\section{Mechanic Robot Design}
\seclabel{hardware_design}

\subsection{Copedo}
\seclabel{copedo}

\mbox{Copedo} was constructed to compete at RoboCup 2012 in Mexico and played in the TeenSize class until 2015.
Initially, Copedo was 114\,cm tall and had a weight of ~8\,kg \cite{missura2013robocup}. To comply with the AdultSize class rules, our robot 
was extended to be 131\,cm tall and 10.1\,kg in weight. The upgrade of 
Copedo was based on another one of our robots---Dynaped, as it was upgraded in 2016~\cite{Farazi2017}.
Copedo has been fitted with new electronics and onboard PC to use our open source ROS-based 
framework just like Dynaped, and our other robots. Copedo is constructed from milled 
carbon fiber parts that are assembled to rectangular shaped legs and flat arms. The torso is 
constructed entirely from aluminum and consists of a cylindrical tube and a rectangular cage that holds 
the main electronic components. During the upgrade process, the cylindrical tube was replaced with a longer one, while the cage
was that of the \nop. To comply with the vision system of our other robots, Copedo received a new 3D printed head,
similar to the one used in the \iguhop~\cite{Allgeuer2015b}. The process and result of the hardware modifications can be seen in \figref{copedo_upgrade}.

\begin{figure}
\parbox{\linewidth}{\centering
\includegraphics[height=47mm]{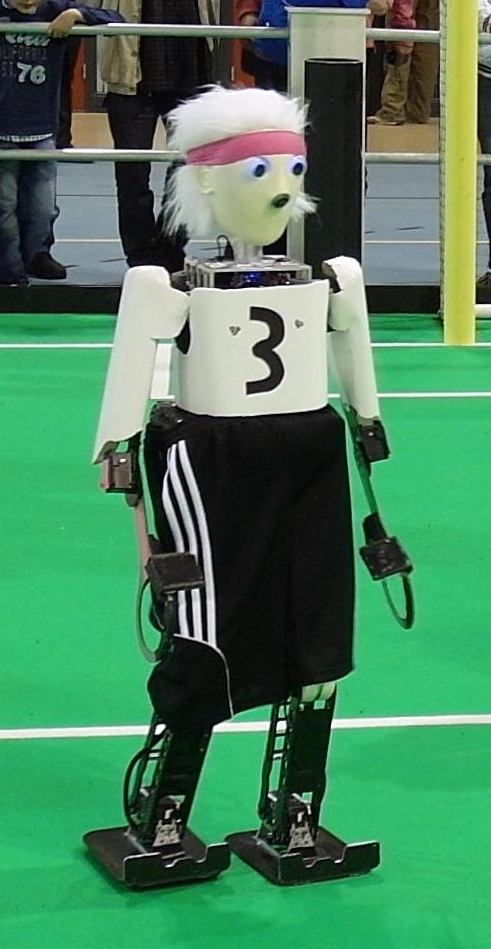}
\hspace{0.005\linewidth}\includegraphics[height=47mm]{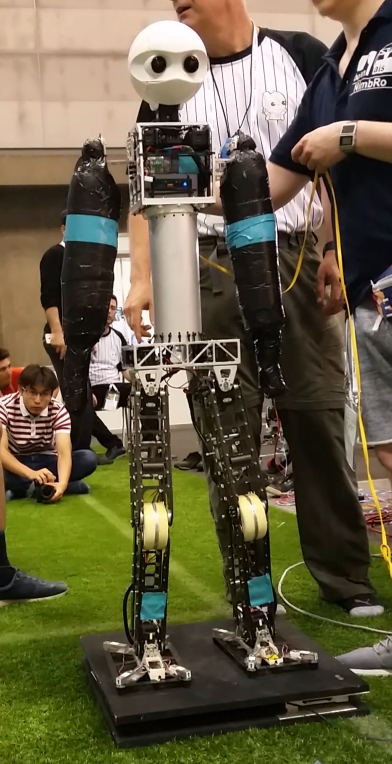}
\hspace{0.005\linewidth}\includegraphics[height=47mm]{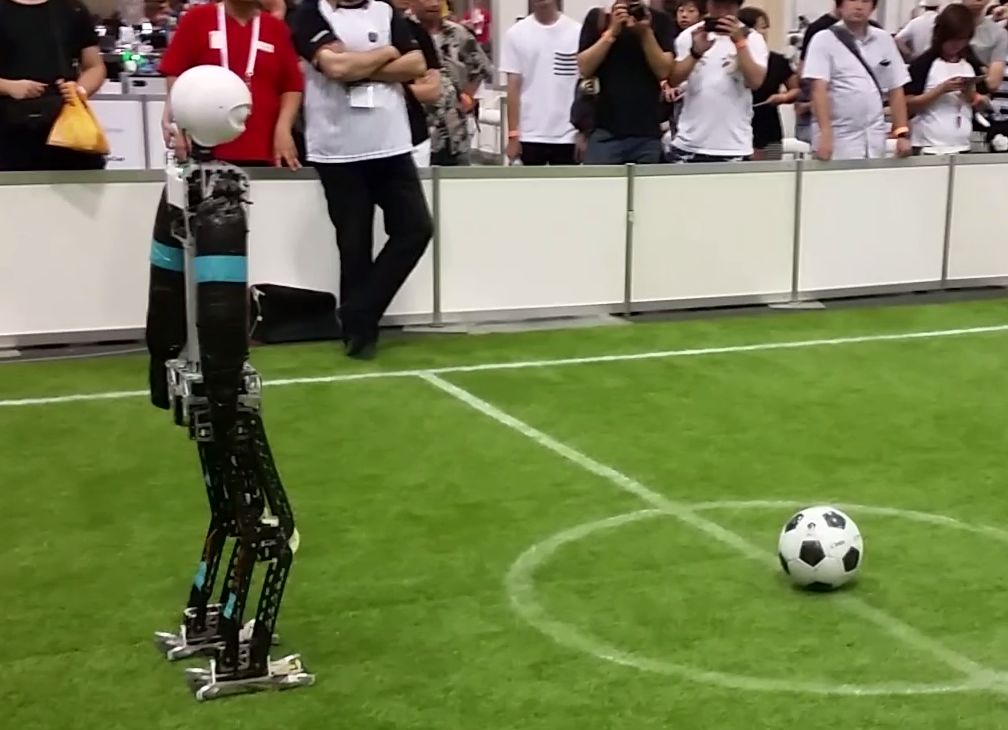}}
\caption{Left: Copedo before the upgrade.
Middle: Copedo after the upgrade. Right: Copedo during the semi-finals at RoboCup 2017 in Nagoya.}
\figlabel{copedo_upgrade}
\vspace{-2ex}
\end{figure}

With a new head allowing for pan and tilt motions, \mbox{Copedo} has 14 actuated
degrees of freedom (DoF). The hip roll, hip pitch, and knee DoF are driven by master-slave pairs of
Robotis Dynamixel EX-106+ actuators. All other DoFs are driven by single actuators
including EX-106+ actuators for ankle roll, EX-106+ actuators for hip yaw and 
RX-64 actuators for shoulder pitch, as well as the neck yaw and pitch.
The robot has been fitted with cleats in the corners of its feet, to assist 
walking on artificial grass. More details on the robot's core mechanical design features can be found in \cite{missura2013robocup}.

Along with the necessary structural upgrade, Copedo has received a new Intel NUC computer with an Intel Core~i7-7567U processor 
operating at a maximum frequency of \SI{4.0}{\giga\hertz}. The PC is fitted with \SI{4}{GB} 
of RAM and a \SI{128}{GB} solid state disk. Available communication 
interfaces include USB 3.0, HDMI, Mini DisplayPort and Gigabit Ethernet. The PC 
is connected to a Robotis CM740 board, which communicates with all actuators 
on a RS485 star topology bus. The CM740 incorporates a 3-axis accelerometer and 
gyroscope for a total of 6 axes of inertial sensory data.

\subsection{NimbRo-OP2}
\seclabel{nop2}

According to the requirements of the RoboCup 2017 Humanoid league, 
we developed a new platform: the \noptwo \cite{ficht2017nop2}. Being almost 135\,cm tall and only 18\,kg in weight, 
the robot is able to participate in both TeenSize and AdultSize classes. The robot's exoskeleton was 3D printed
with an industrial-grade SLS printer out of PA-12 nylon, which contributes greatly to its low weight.
In terms of electronics, the robot uses the same Intel NUC in combination with the Robotis CM740 as Copedo. 
There also exists the possibility of further upgrading the unit, through the inclusion of standard VESA75 and VESA100 mounts.

The structure of the robot has been simplified as much as possible to maintain low complexity, 
but retain functionality that is required when playing soccer. The robot in its entirety uses 34 Dynamixel MX-106 actuators. 
The upper body kinematics consists of three serial chains---two arms and a neck connecting the head to the trunk. 
The neck consists of a yaw and a pitch joint, while the arms have two pitch and one roll joint.
In the legs, we decided to use parallel kinematics along with external gearing to allow for more torque output, which 
is necessary for dynamic walking. 

The design made use of our long experience with 
building humanoid robots. Creating a new robot is a complex and time-consuming process, of which 
manufacturing plays a big role. By taking advantage of the versatility of 3D printing, we were
able to develop an affordable, customisable, highly-capable, adult-sized humanoid robot in little time.
The whole design and manufacturing process took less than six months. A view of the finished robot and some of its key design
features can be seen in \figref{op2_design}.

\begin{figure}[!t]
\parbox{\linewidth}{\centering
\includegraphics[width=1.0\linewidth]{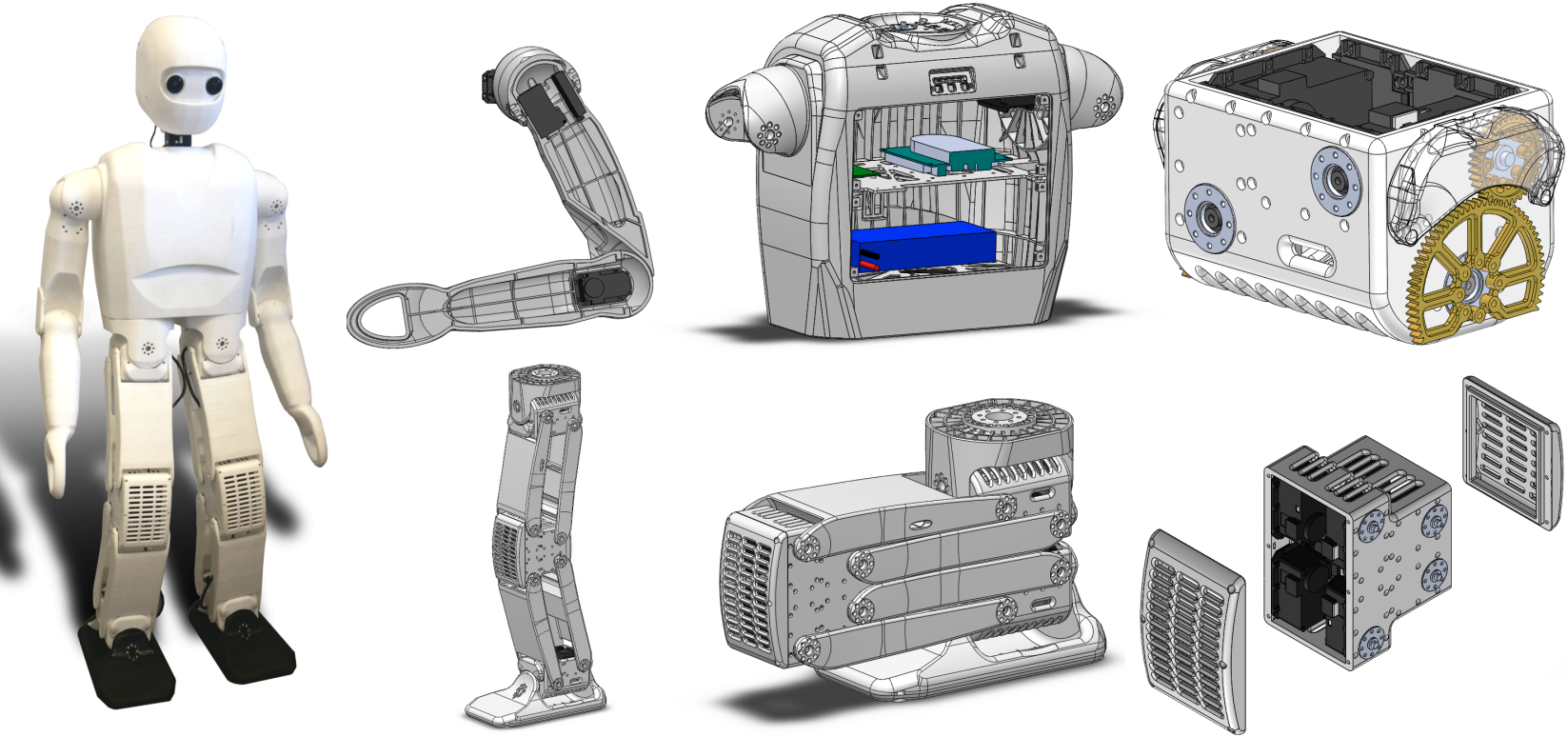}}
\caption{Left: Assembled NimbRo-OP2. Right: CAD visualisation of key robot components: extended and compressed parallel kinematics leg, cross section of the arm, rear view on the trunk, knee and ankle/hip joints.}
\figlabel{op2_design}
\vspace{-2ex}
\end{figure}

Simplicity was one of the key points in creating the design. This is best shown when looking at the knee of the robot,
which houses eight actuators with only three plastic parts, two of which are covers that are not critical for bearing the load.
Upon removal of the covers, servos are easily accessible for any maintenance that might be needed. The servos themselves 
are exposed through strategic venting hole placement to help with cooling. Most of the parts in the design are symmetrical, meaning that 
a single part is used in multiple spots. This minimises the amount of spare parts needed for repairs, in case a part should break. 
This can be seen in the ankle and hip joints, the knee joints as well as the thigh and shank links.

\section{Software Design}
\seclabel{software_design}

\subsection{Visual Perception}
\seclabel{perception}

Our humanoid robots perceive the environment through digital cameras. 
Each robot is equipped with one Logitech C905 camera, utilising a wide-angle lens 
with an infrared cut-off filter. In this configuration, its field-of-view is nearly 150\degree. 
Our vision system, amongst many other features, can reliably perceive game-relevant objects 
using texture, shape, brightness, and colour information. We project each object into egocentric world 
coordinates by using the intrinsic and extrinsic camera parameters. Variations in hardware result in
projection errors, which scale with distance from the object. We calibrate the position and orientation of the camera frame
with the Nelder-Mead \cite{nelder1965simplex} method. More details on our vision system can be found in \cite{farazi2015}.

\subsubsection{Landmark Detection:}

A number of landmarks can be distinguished on the field, which can be used for localisation.
These include line junctions, goal posts, penalty marks and the center circle.
Field lines are the most useful source of information when it comes to localisation. We detect them 
by using a Canny edge detector on the V channel in the HSV colour space.
We then apply a probabilistic Hough line detector \cite{matas2000robust} to extract line segments
of minimum size to filter out white non-line objects. These segments are then connected to produce 
less, but longer lines, finally achieving a set of lines that correspond to field lines and the center circle.
An example output of our vision system detecting selected objects can be seen in \figref{vision_output}.

\begin{figure}[!t]
\parbox{\linewidth}{\centering
\includegraphics[width=0.45\linewidth]{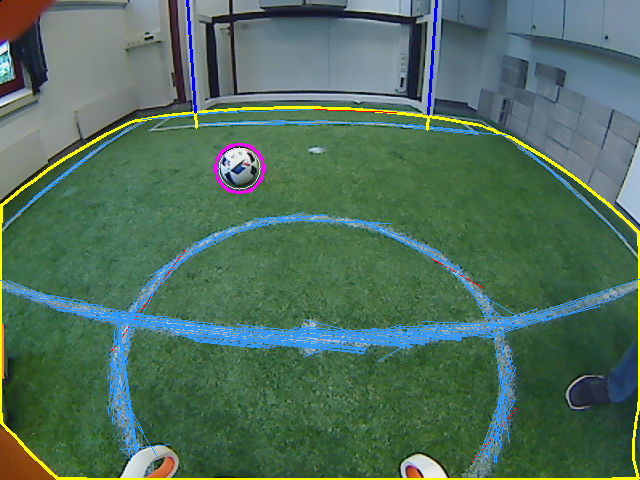}\hspace{0.019\linewidth}\includegraphics[width=0.45\linewidth]{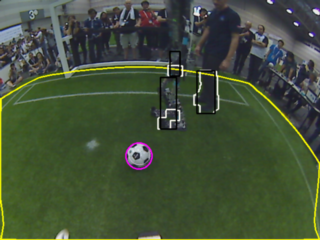}}
\caption{Left: A captured image with detected ball (pink circle), field lines (light blue), field boundary (yellow lines), and goal post (dark blue lines).
Right: Visualisation of our obstacle detection.}
\figlabel{vision_output}
\vspace{-2ex}
\end{figure}

\subsubsection{Localisation and Breaking the Symmetry:}
\seclabel{localisation}

To solve the global localisation problem, our method relies on having a source of 
global yaw rotation of the robot. We used to utilise the compass sensor in previous 
years, but due to the magnetometer ban, introduced in the rules of the Humanoid League in 2017, 
we now use the integrated gyro measurements as the source for yaw orientation. In our case, 
gyro integration is a reliable source of orientation tracking, but it needs a global reference. 
In order to set the initial heading, we could either use manual initialisation or automatic 
initial orientation estimation. Although manual initialisation can be done once before the start 
of each game, it can fail during the match. Sometimes restarting the operating system of the 
robot is unavoidable, which will force a reinitialisation of the heading. As a result, we 
reformulated the heading initialisation problem as a classification task. 

\begin{figure}[!t]
\parbox{\linewidth}{\centering
\includegraphics[width=0.69\linewidth]{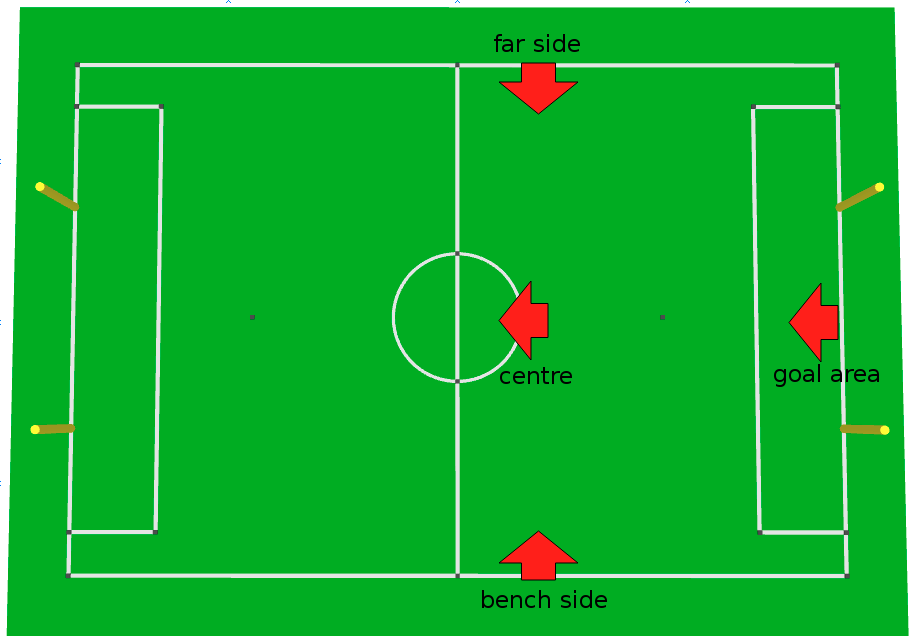}}
\caption{Set of predefined positions the robot can start in.}
\figlabel{init_poses}
\vspace{-2ex}
\end{figure}

According to the rules, there are a few predefined positions and orientations that the robot can start in or 
enter the game from. As shown in \figref{init_poses}, the robot can start in four different
positions. In two of the spots, it should face the opponent goal---near the center circle 
and goal area. The other two sets of locations are at the sideline in the robot's own half---facing
the field. We employ a multi-hypothesis version of our localisation module, which is initialised with 
four instances of initial hypothetical locations. During a brief period at the beginning, 
the robot tries to find the most probable hypothesis among all running instances. This
stops when either the process times out or the robot finds the best hypothesis. Finally, 
the vision module keeps the best instance and discards the rest. To make sure that the 
decision is correct, we double check the result based on the perceived landmarks like 
goalposts and the center circle.

\subsubsection{Obstacle Detection:}
\seclabel{obstacles}

Obstacle detection is a crucial ability in the game, especially when the robot 
is handling the ball. In our software, obstacle detection is done mainly based on a 
model of colour distribution on the perceived robots. By having the minimum and maximum 
height of the robot in each size class, we roughly know what size to expect in each 
distance from the observer. We search for the respected bounding box size in each 
distance level and discard obstacle candidates that are not in the expected size range. 
After detecting each obstacle bounding box, we compare the colour histogram of each of the 
bounding boxes to the expected model of the obstacle, which are then labeled as 
either teammate, rival robot, or the referee. The detection history is then clustered in 
egocentric world coordinates and filtered based on the location of each cluster. Finally, 
to make the output robust against false negatives, we predict the expected movement of the obstacle
in accordance to the estimated changes in the robot's location. Each of the clusters 
has a certainty level, which is increased when it is detected, and decreased otherwise. The effect 
of this method can be seen in \figref{vision_output}.

\subsection{Soccer Behaviours}

\begin{figure}[!t]
\parbox{\linewidth}{\centering%
\includegraphics[width=0.49\linewidth]{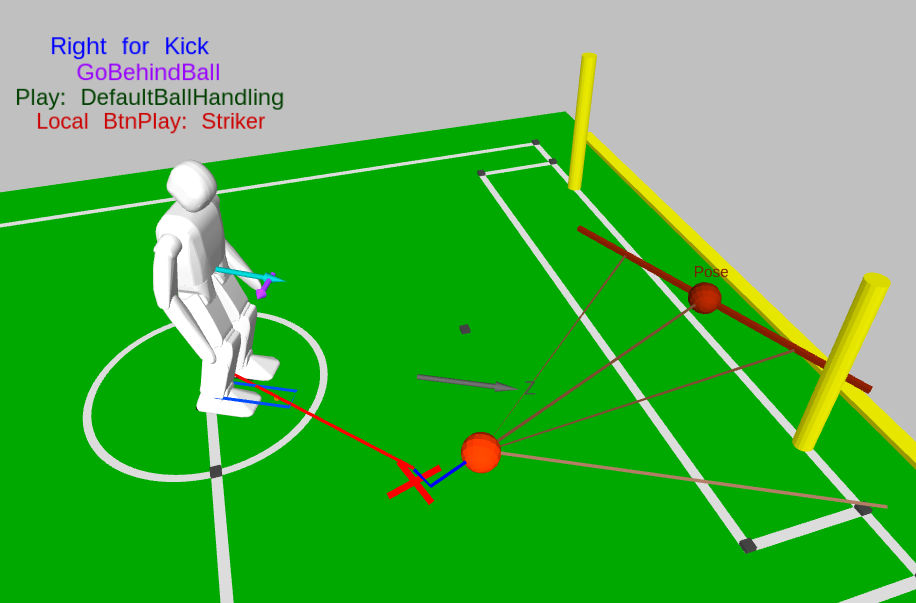}\hspace{0.01\linewidth}%
\includegraphics[width=0.49\linewidth]{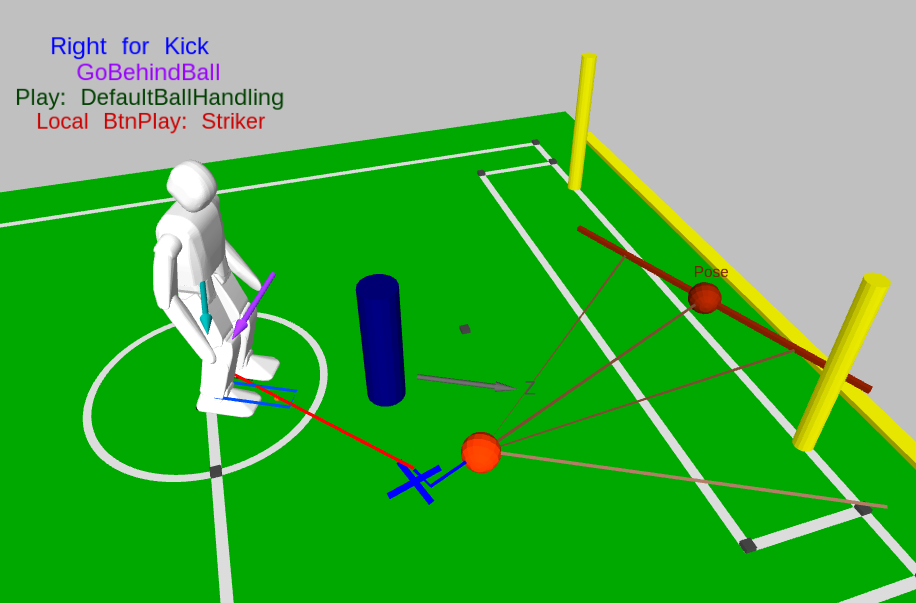}\\[1mm]
\includegraphics[width=0.49\linewidth]{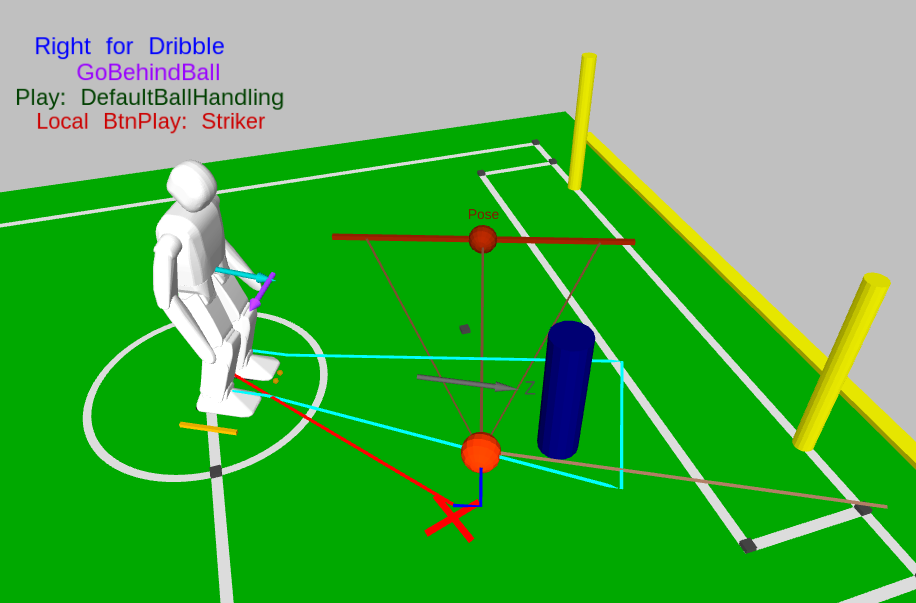}\hspace{0.01\linewidth}%
\includegraphics[width=0.49\linewidth]{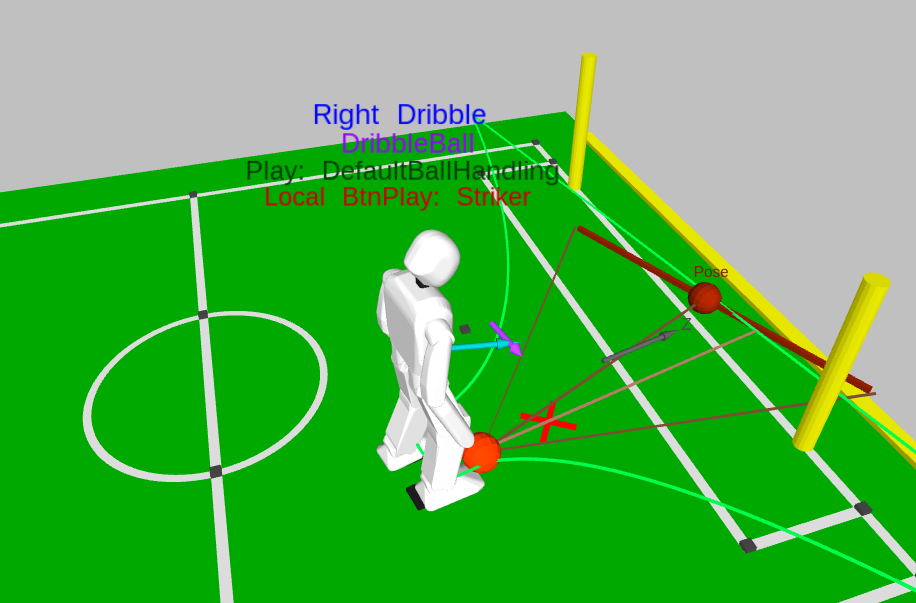}}
\caption{Simulation screenshots showing the behaviour visualisations for the {Go 
Behind Ball} behaviour (top left), how this changes when an obstacle is blocking 
the ball (top right), when an obstacle is blocking the path from 
the ball to the goal (bottom left), and the dribble behaviour (bottom right).}
\figlabel{behaviours}
\end{figure}

Once the visual perception module has established the current state of the game, 
including the localised pose of the robot on the field and the positions of the 
ball and any obstacles, the behaviour module uses this information to control 
the soccer actions of the robot. Two finite state machines (FSM) have been 
implemented in a hierarchical fashion on top of one another, to process the 
available data and output the required walking velocities and motion triggers 
for the robot. The higher-level FSM, called the Game FSM, is responsible for 
deciding on game-level actions, such as whether to try to score a goal, or 
perform auto-positioning, or defend the goals. The lower-level FSM on the other 
hand, called the Behaviour FSM, implements fundamental soccer skills such as 
dribbling and kicking the ball, and walking to a localised ball position while 
avoiding obstacles.

\subsubsection{Ball Approach:}
Approaching the ball is a fundamental skill on the level of the Behaviour FSM. 
The robot should walk as efficiently as possible to the required position behind 
the ball, without pushing the ball away. A circular halo is established around 
the ball, and the approach phase is divided into the near and far cases. In the 
far case, the robot only turns and walks forward, to maximise its speed and 
stability in covering ground, while in the near case, slower side-stepping is 
added to allow the robot to get around the ball efficiently. This is illustrated 
on the top left in \figref{behaviours}.

\subsubsection{Obstacle Avoidance:}
If an obstacle blocks the approach to the ball, obstacle avoidance is 
applied on the level of the Behaviour FSM. The robot is slowed down, yaw rotation is 
added to turn it away from the obstacle, and the linear velocity of the robot is 
rotated to limit the radial component of the velocity towards the obstacle (see 
top right in \figref{behaviours}). The maximum allowed radial velocity is a 
function of the proximity of the obstacle, and can also be negative, pushing the 
robot away from the obstacle.

\subsubsection{Obstacle Ball Handling:}
If an obstacle is close to the ball or blocks the path from the ball to the 
goals, obstacle ball handling is applied on the level of the Game FSM. This 
rotates the ball target away from obstructions, and if the ball target is no 
longer safely in goals, or the obstacle is too close to the ball, dribbling is 
enforced for safety, as shown on the bottom left in \figref{behaviours}. The 
correct foot to dribble with is also enforced to avoid collisions with the 
obstacle as much as possible.

\subsubsection{Kicking and Dribbling:}
When the robot is behind the ball and aligned with the ball target, either the 
{Kick or Dribble} behaviours activate as required, and drive the ball towards the 
target. When {Dribble} activates (see bottom right in \figref{behaviours}), a 
large variance of ball position is tolerated that allows dribbling to continue. 
The robot however consistently tries to correct for any misalignments in its 
dribble approach. If the dribble lock is lost, the ball approach is started once 
more.

\section{Performance in Soccer Tournament}
\seclabel{performance}

At RoboCup 2017, eight teams participated in the soccer tournament.
Our robot NimbRo-OP2 played three round robin games vs. teams ZSTT (Taiwan \& Korea), KIS (Japan), and CIT Brains (Japan). 
The robot scored very reliably, such that the three games were clearly won with a total score of 26:0.
In semi-final, our robots NimbRo-OP2 and Copedo played vs. team IRC (IRC) and clearly won with a score of 9:0.

In the final, our robot NimbRo-OP2 faced the robot Sweaty (Germany).
Sweaty had shown good walking and kicking capabilities during the competition.
The two robots, fighting for the ball are shown in \figref{finals}.
Our robot had clear advantages in such fights, avoiding the opponent and maintaining balance while Sweaty often lost balance and had to leave the field. NimbRo-OP2 scored reliably. After 7:0 at half time, the game ended early with a score of 11:1.

During the competition games, \noptwo has shown a very stable and fast walk.
It never lost its balance when walking in free space.
The only occasions which would lead to the fall were situations when a strong collision with other robots occurred.
The kicking was robust and strong, allowing us to score far goals.
Finally, the localisation worked reliably during all games and obstacle avoidance allowed to win fights for the ball by dribbling it around the opponent, which was often the case in the finals.

\begin{figure}[!t]
\centering
\includegraphics[height=0.32\linewidth]{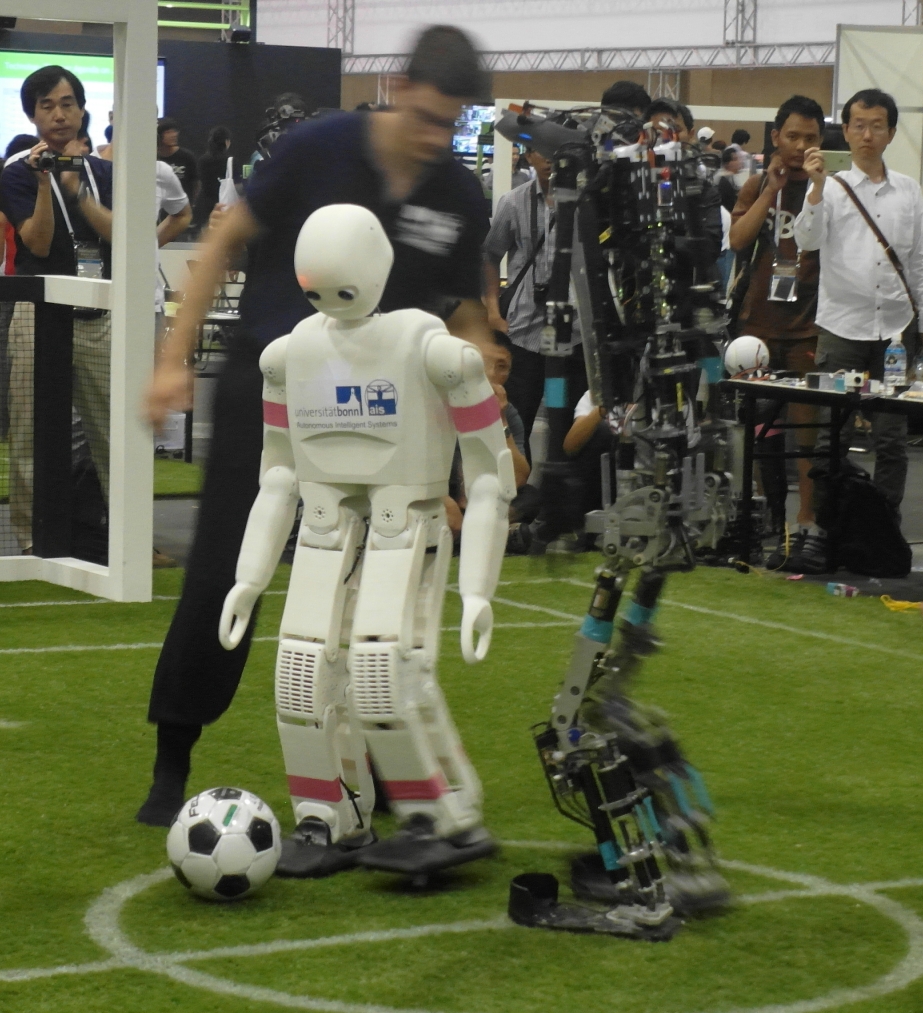}~
\includegraphics[height=0.32\linewidth]{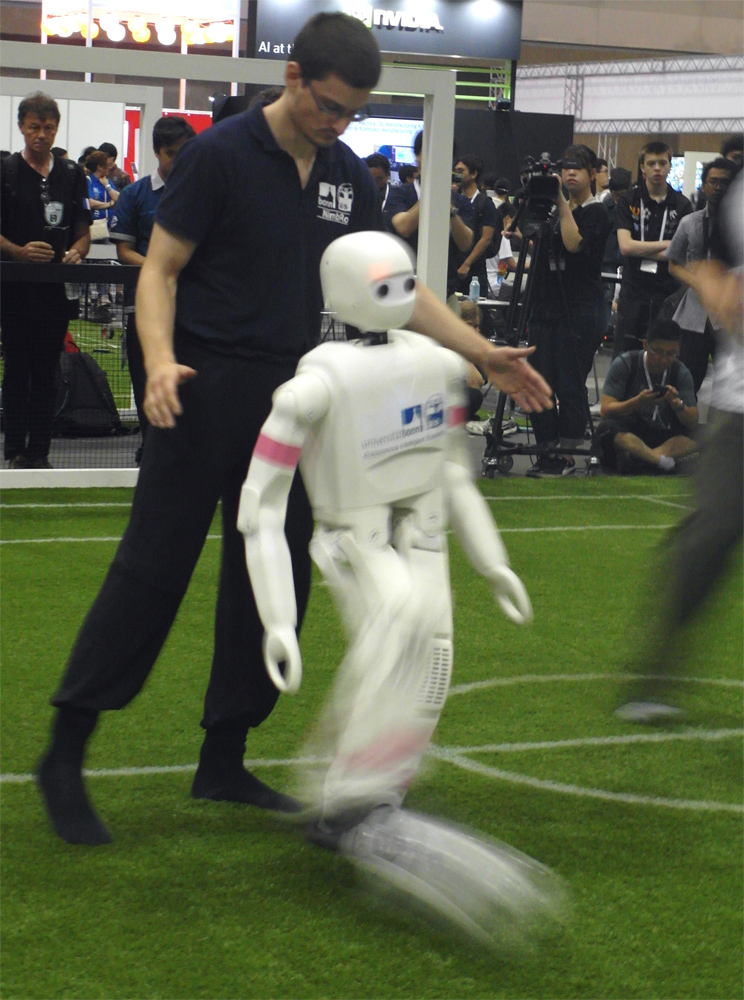}~
\includegraphics[height=0.32\linewidth]{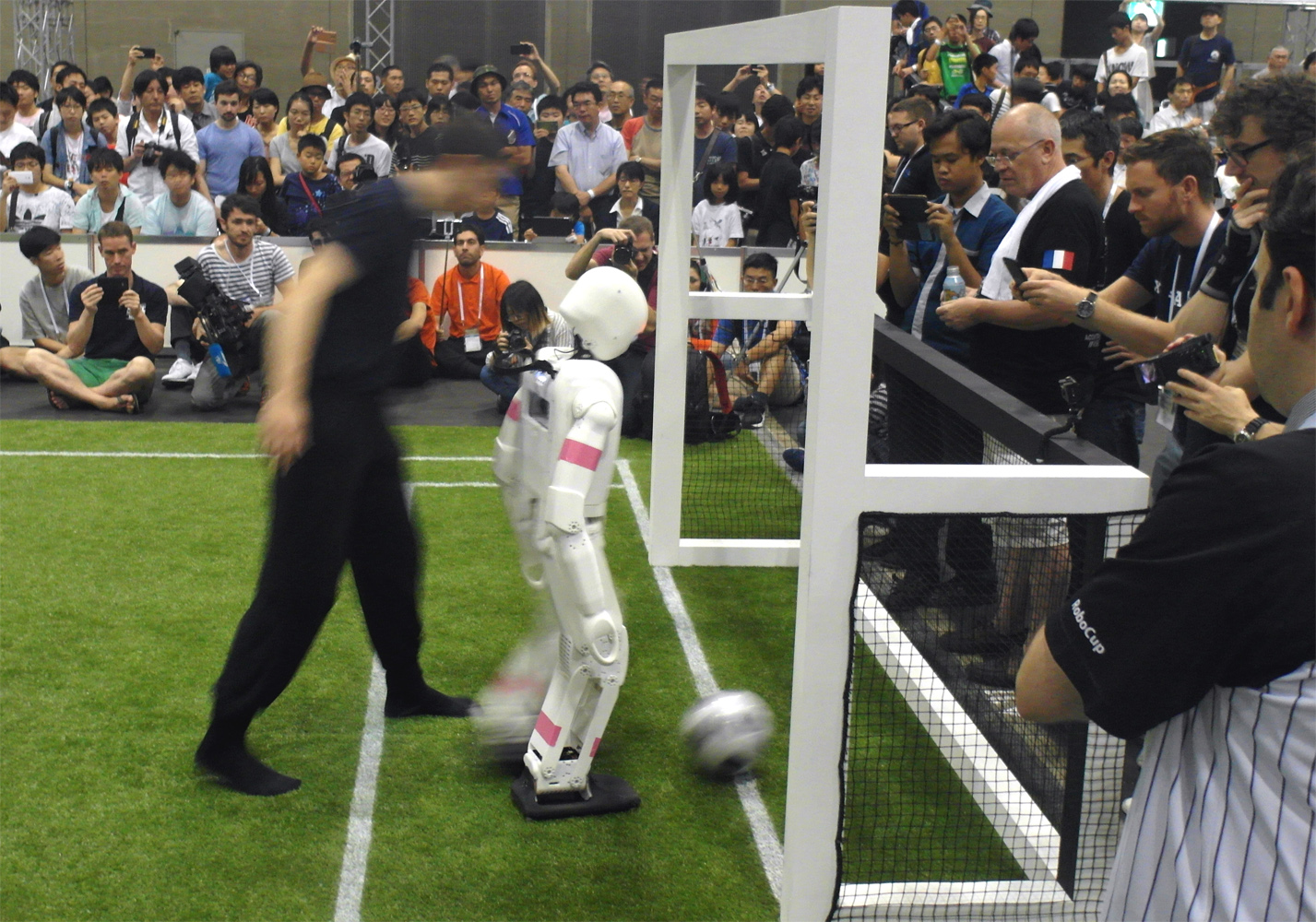}\vspace*{-1ex}
\caption{Impressions from the RoboCup 2017 AdultSize final vs. Sweaty (Germany).}
\figlabel{finals}
\end{figure}

\section{Technical Challenges}
\seclabel{technical_challenges}

In addition to the main competition, technical challenges test how well a robot can perform a specific task in isolation.
At RoboCup 2017, four technical challenges were posed: Push Recovery, High Jump, High Kick, and Kick a Moving Ball.
In this section we present our strategy for three technical challenges which were addressed for RoboCup 2017.

\subsection{Push Recovery}
\seclabel{push_recovery}

The goal of this challenge is to withstand a strong push.
An impact is applied to the robot on the level of the CoM by a pendulum.
To define the impulse, the 3\,kg weight is retracted by a distance $d$ from the point of contact with the robot.
The push is applied only from the front and from the back.
The robot has to be walking on the spot during the whole challenge.

During the challenge we used the gait~\cite{Allgeuer2016a} completely unchanged from the configuration in the regular games.
\noptwo was able to successfully withstand a push with $d=70$\,cm.
Our robot, recovering after the push is shown in \figref{push_recovery_and_jump}.

\begin{figure}[!t]
\parbox{\linewidth}{\centering
\includegraphics[height=58mm]{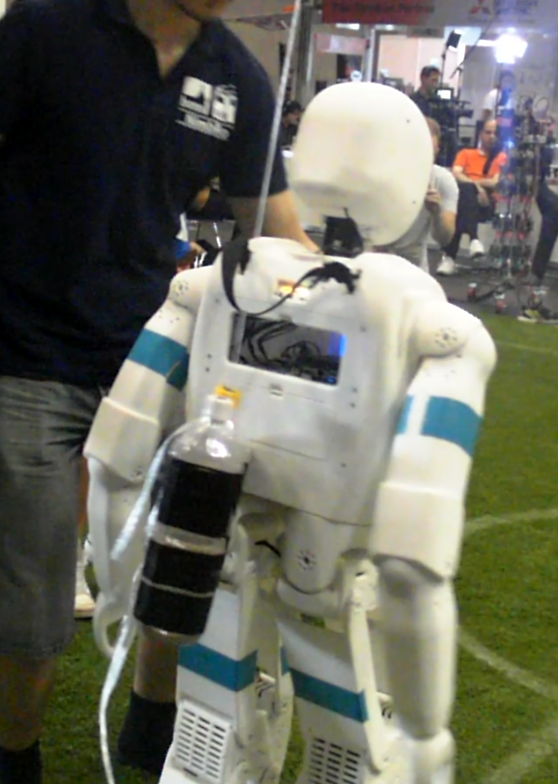}
\hspace{0.019\linewidth}\includegraphics[height=58mm]{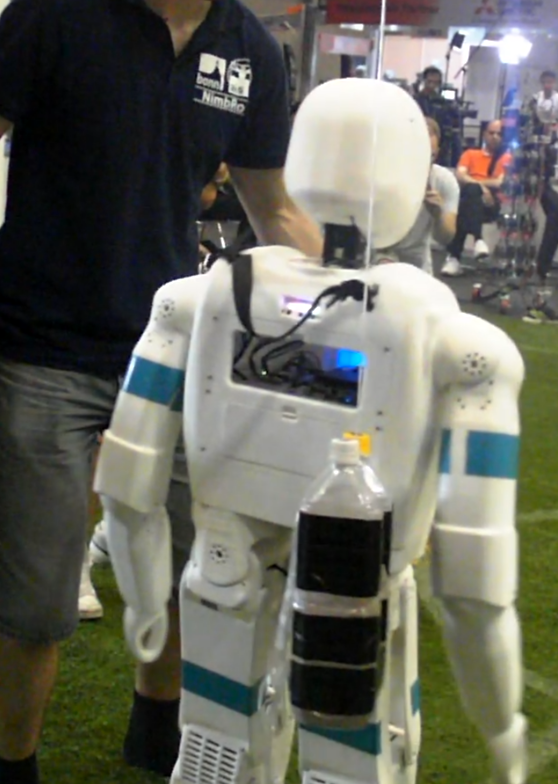}
\hspace{0.019\linewidth}\includegraphics[height=58mm]{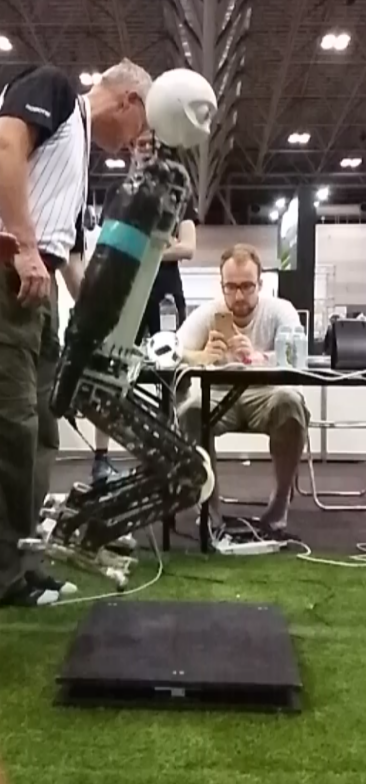}}
\caption{Our robots performing the technical challenge.
Left, Middle: \noptwo performing push recovery. 
Right: Copedo performing the high jump.}
\figlabel{push_recovery_and_jump}
\vspace{-2ex}
\end{figure}

\subsection{High Jump}
\seclabel{high_jump}

The goal of high jump is to jump as high as possible and remain airborne as long as possible.
The challenge was performed using a predesigned jump motion, which was constructed with a keyframe editor.
Execution of this motion on Copedo resulted in a partially successful trial.
The robot jumped higher than 10\,cm, but upon landing a stable posture was not held.
Copedo performing the jump is depicted in \figref{push_recovery_and_jump}.

\subsection{Goal Kick from Moving Ball}
\seclabel{moving_ball}

This challenge requires a robot to score a goal by kicking a moving ball into the goal.
The ball is rolled from a ramp along the goal area line.
The difficulty is defined by the angle of the ramp, and the distance $d_{ramp}$ from the endpoint of the ramp to the ball in its initial static position.
We solved for this task is as follows:
\begin{enumerate}
\item Before the ball is released, execute a pre-kick motion. 
During this motion, the robot goes from a standard standing posture to a posture with one support leg.
The other leg is lifted and folded in order to be ready for kicking.
\item Estimate the time $t_{arrive}$ needed for the ball to arrive in the region where it can be kicked into the goal by the previously folded leg.
\item Execute the kicking motion when the ball is in the kick region according to the estimation from the previous step.
\end{enumerate}

The motions described above are designed using our keyframe editor.
In order to estimate the time needed for the ball to arrive in the kick region, 
we utilise the ball detection information obtained from the vision module, described in \secref{perception}.
Ball observations arrive in real-time and are stored in a stack $S$, where $\forall s \in S: s= \langle p,r,t \rangle$, where $p$ is a confidence of the detection,
$r$ is the estimated position of the ball, and $t$ is a time stamp of the measurement.
The position $r$ is represented in a 2D coordinate frame which covers the surface of the playing field and has its origin at the position of the robot.

We estimate the velocity of the ball with the two latest measurements $s_1, s_2: p_1, p_2 > p_{min} \vee t_2-t_1 > \delta t$,
where $p_{min}$ is a predefined minimal confidence for the measurement, and $\delta t > 0$ is a predefined minimal time difference between two measurements.
Given such measurements $s_1$ and $s_2$, we estimate the velocity $v$:
\begin{equation}\label{eq:ball_velocity}
v = \frac{d(r_2, r_1)}{t_2-t_1},
\end{equation}
where $d(\cdot,\cdot)$ is the Euclidean distance.
Each obtained estimate $v$ is pushed into a double-ended queue $V$ of maximum size $N$.
We smooth the velocity by averaging over $V$:
\begin{equation}\label{eq:ball_velocity_avg}
v_{smooth} = \frac{1}{|V|}\sum_{i=1}^{N}V_{i}.
\end{equation}
Finally, we estimate the time of arrival of the ball using the smoothed velocity and the most recent pair of measurements:
\begin{equation}\label{eq:time_arrival}
t_{arrive} = \frac{d(r_{kick}, r_2)}{v_{smooth}},
\end{equation}
where $r_{kick}$ is a predefined position of the ball which is optimal for performing the kick.

This simple strategy worked well for this challenge, because the ball moves with a velocity that is close to uniform.
In addition, the region where the ball can be kicked into the goal is by far not limited to the position $r_{kick}$.
That is why small deviations of the estimate $t_{arrive}$ from the ground truth do not cause failures.
At RoboCup 2017, we used the following parameters: $p_{min}=0.5$, $\delta t=0.1$ and $N=3$.
\noptwo was able to score a goal from a moving ball with the largest possible $d_{ramp}$, as shown in \figref{moving_ball}.

\begin{figure}[!t]
\parbox{\linewidth}{\centering
\includegraphics[width=0.3\linewidth]{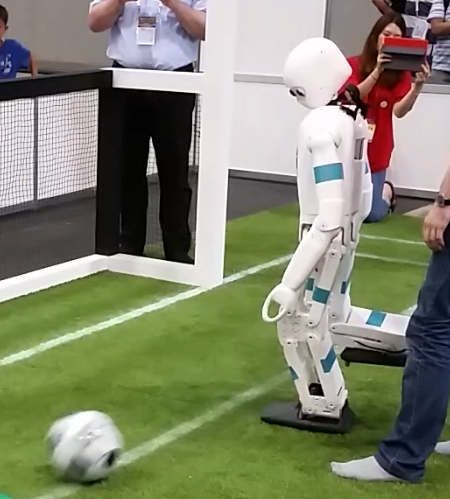}
\hspace{0.019\linewidth}
\includegraphics[width=0.3\linewidth]{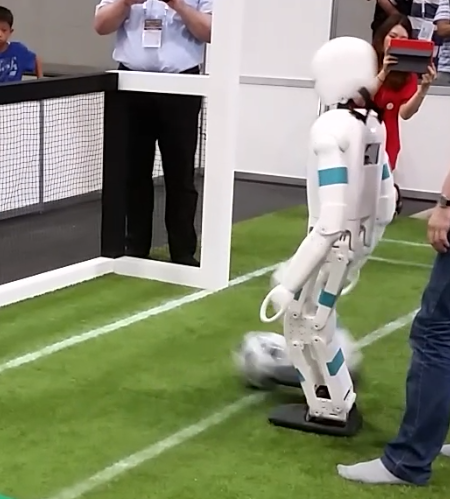}
\hspace{0.019\linewidth}
\includegraphics[width=0.3\linewidth]{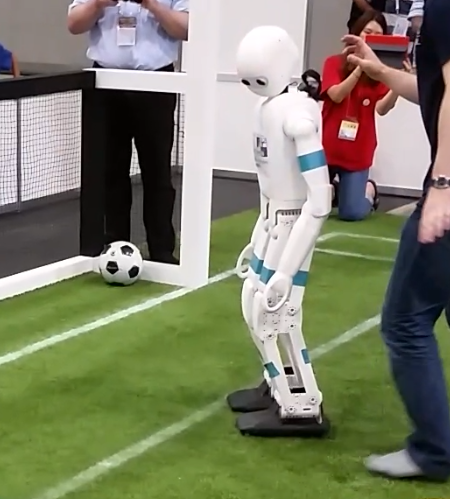}}
\caption{\noptwo kicking a moving ball. 
Left: The robot is waiting for the ball in pre-kick posture with right leg prepared for a kick. 
Middle: The ball reached the target location and the kick is being performed.
Right: Goal successfully scored, stable posture is reached.}
\figlabel{moving_ball}
\vspace{-2ex}
\end{figure}

\section{Conclusions}

In this paper, we described the mechatronic design of our robots and some of the perception and control 
approaches that led to our success in the Humanoid League AdultSize soccer competition and 
technical challenges. At RoboCup 2017, we participated in the AdultSize category for the first time
and with very little time to prepare, we were able to produce two robots that were able to secure
a victory in both competitions. Over a span of five soccer games we aggregated a total score of 46:1, conceding
a single goal in the finals, and gathered 21 points from three technical challenges. A video showing the competition highlights is available online\footnote{RoboCup 2017 NimbRo AdultSize highlights: \url{https://youtu.be/RG205OwGdSg}}.
Our new robot NimbRo-OP2 was also 
awarded the RoboCup Design Award by Flower Robotics. The hardware of the NimbRo-OP2\footnote{Hardware: 
\url{https://github.com/NimbRo/nimbro-op2}} as well as our software\footnote{Software: \url{https://github.com/AIS-Bonn/humanoid_op_ros}} 
were released open-source to GitHub with the hope that other teams and research groups benefit from our work.

\section{Acknowledgements}
\footnotesize
This work was partially funded by grant BE 2556/13 of the German Research Foundation (DFG).

\bibliographystyle{ieeetr}
\bibliography{ms}

\begin{thebibliography}{10}

\bibitem{Lee:RoboCup2011}
D.~D. Lee, S.~Yi, S.~G. McGill, Y.~Zhang, S.~Behnke, M.~Missura, H.~Schulz,
  D.~W. Hong, J.~Han, and M.~A. Hopkins, ``{RoboCup} 2011 {Humanoid} {League}
  winners,'' in {\em RoboCup 2011: Robot Soccer World Cup {XV}}, pp.~37--50,
  Springer, 2011.

\bibitem{missura2013robocup}
M.~Missura, C.~M{\"u}nstermann, M.~Mauelshagen, M.~Schreiber, and S.~Behnke,
  ``Robo{C}up 2012 {B}est {H}umanoid {A}ward {W}inner {N}imb{R}o
  {T}een{S}ize,'' in {\em RoboCup 2012: Robot Soccer World Cup XVI},
  pp.~89--93, Springer, 2013.

\bibitem{Missura:RoboCup2013}
M.~Missura, C.~M{\"{u}}nstermann, P.~Allgeuer, M.~Schwarz, J.~Pastrana,
  S.~Sch{\"{u}}ller, M.~Schreiber, and S.~Behnke, ``Learning to improve capture
  steps for disturbance rejection in humanoid soccer,'' in {\em RoboCup 2013:
  Robot World Cup {XVII}}, pp.~56--67, Springer, 2013.

\bibitem{Farazi2017}
H.~Farazi, P.~Allgeuer, G.~Ficht, A.~Brandenburger, D.~Pavlichenko,
  M.~Schreiber, and S.~Behnke, {\em {RoboCup} 2016 {Humanoid} {TeenSize} Winner
  {NimbRo}: {Robust} Visual Perception and Soccer Behaviors}, pp.~478--490.
\newblock Springer, 2017.

\bibitem{Allgeuer2015b}
P.~Allgeuer, H.~Farazi, M.~Schreiber, and S.~Behnke, ``{Child-sized 3D Printed
  igus Humanoid Open Platform},'' in {\em Proceedings of 15th IEEE-RAS Int.
  Conference on Humanoid Robots (Humanoids)}, 2015.

\bibitem{ficht2017nop2}
G.~Ficht, P.~Allgeuer, H.~Farazi, and S.~Behnke, ``Nimb{R}o-{O}{P}2: {G}rown-up
  3{D} {P}rinted {O}pen {H}umanoid {P}latform for {R}esearch,'' in {\em
  Proceedings of 17th IEEE-RAS Int. Conference on Humanoid Robots (Humanoids)},
  2017.

\bibitem{nelder1965simplex}
J.~A. Nelder and R.~Mead, ``A simplex method for function minimization,'' {\em
  The Computer Journal}, vol.~7, no.~4, pp.~308--313, 1965.

\bibitem{farazi2015}
H.~Farazi, P.~Allgeuer, and S.~Behnke, ``A {M}onocular {V}ision {S}ystem for
  {P}laying {S}occer in {L}ow {C}olor {I}nformation {E}nvironments,'' in {\em
  10th Workshop on Humanoid Soccer Robots, IEEE-RAS Int. Conference on Humanoid
  Robots}, 2015.

\bibitem{matas2000robust}
J.~Matas, C.~Galambos, and J.~Kittler, ``Robust detection of lines using the
  progressive probabilistic {H}ough transform,'' {\em Vision and Image
  Understanding}, 2000.

\bibitem{Allgeuer2016a}
P.~Allgeuer and S.~Behnke, ``Omnidirectional {B}ipedal {W}alking with {D}irect
  {F}used {A}ngle {F}eedback {M}echanisms,'' in {\em Proceedings of 16th
  IEEE-RAS Int. Conference on Humanoid Robots (Humanoids)}, (Canc\'un, Mexico),
  2016.

\end{thebibliography}

\end{document}